# Using Natural Language for Human-Robot Collaboration in the Real World

**Peter Lindes[1], Kaoutar Skiker[2]**

[1]Center for Integrated Cognition  [2]Conitive
peter.lindes@cic.iqmri.org  kaoutar.skiker@conitive.com

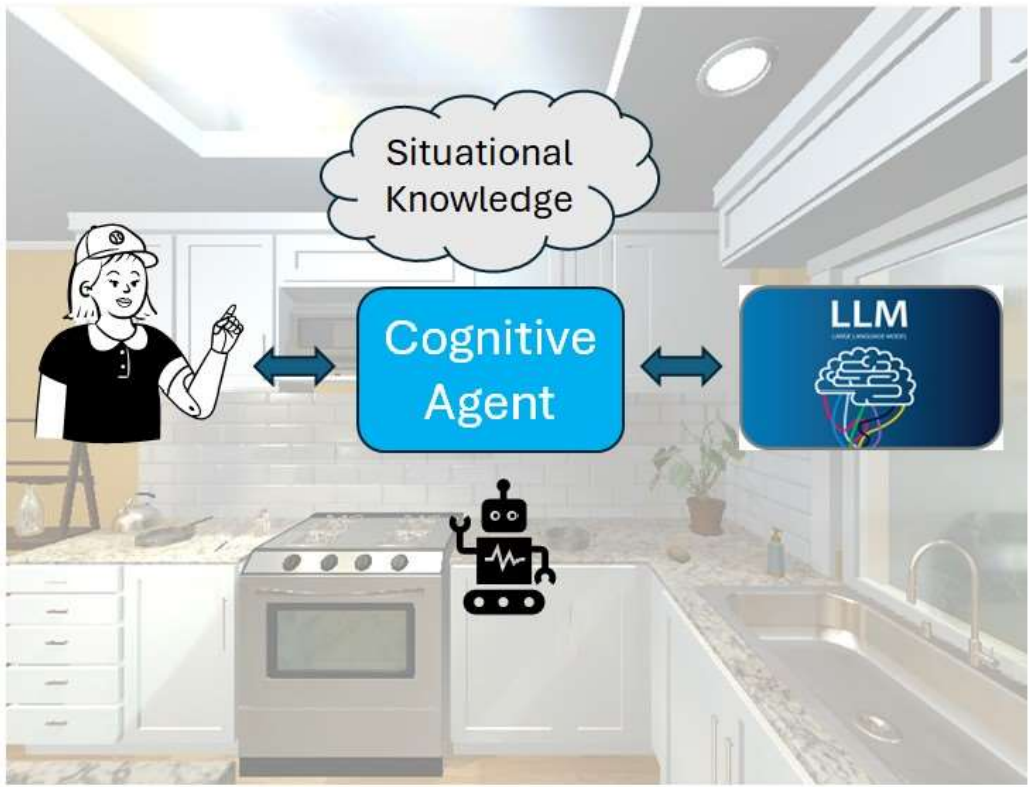

*Figure 1: A Collaborative System that combines the memory and reasoning capabilities of a cognitive agent that controls a robot with its abilities to interact with a human director using language and to make requests to an LLM to help it understand the human's natural language.*

## Abstract

We have a vision of a day when autonomous robots can collaborate with humans as assistants in performing complex tasks in the physical world. This vision includes that the robots will have the ability to communicate with their human collaborators using language that is natural to the humans. Traditional Interactive Task Learning (ITL) systems have some of this ability, but the language they can understand is very limited. The advent of large language models (LLMs) provides an opportunity to greatly improve the language understanding of robots, yet integrating the language abilities of LLMs with robots that operate in the real physical world is a challenging problem.

In this chapter we first review briefly a few commercial robot products that work closely with humans, and discuss how they could be much better collaborators with robust language abilities. We then explore how an AI system with a cognitive agent that controls a physical robot at its core, interacts with both a human and an LLM, and accumulates situational knowledge through its experiences, can be a possible approach to reach that vision. We focus on three specific challenges of having the robot understand natural language, and present a simple proof-of-concept experiment using ChatGPT for each. Finally, we discuss what it will take to turn these simple experiments into an operational system where LLM-assisted language understanding is a part of an integrated robotic assistant that uses language to collaborate with humans.

# 1. Introduction

Today's robotic technology includes many commercial robots that perform important tasks in the real world. In some cases, they collaborate with humans, but usually their ability to collaborate is limited (Bansal et al., 2024). One of the biggest limitations is a lack of the ability to understand natural language that is both grounded in the immediate situation and truly natural for the human collaborator to use (Tellex et al. 2020). The purpose of this chapter is to explore a possible path toward using natural language to work with a robot in the real world.

Figure 1 gives an overview of the sort of system we are considering. At the center is a cognitive agent that is the main brain of this artificial agent. It can interact with a human using natural language, and it receives perception from and controls the actions of a physical robot. It contains a large store of situational knowledge, most of which it learns gradually over time through its experiences. It also has a connection to a large language model (LLM; Wu et al., 2023). The agent makes requests to the LLM to translate human language into simpler or formal forms that it can understand reliably, and also to ask for increments of general or common-sense knowledge. As a working example, we focus on an agent intended to be a private chef in a home kitchen that will, over time, learn to be more and more capable of performing complex tasks, like *Fix my breakfast.* or *Make dinner.* Overall, the integrated system should be capable of learning through collaboration using natural language to perform such tasks.

Toward this goal, we present some examples of currently available commercial robots that work in close proximity to humans. Then we describe a complex agent architecture that we believe could be capable of providing collaborative use of language. Next, we explore three major challenges in developing this language capability, presenting a simple LLM experiment for each. Finally, we suggest a road map to achieving a robust language capability.

# 2. Collaborative Robots in the Real World

At the outset we want to consider some of the many ways that robots are currently used in human-centered situations. In this section we describe several commercially available robots for a variety of applications. In each case we give a link to where this information was obtained in July 2025.

## 2.1 Moxi: Human-Robot Collaboration in a Hospital

Diligent Robotics makes a robot called Moxi that works as an assistant to nurses and other staff in hospitals. They express their vision this way:

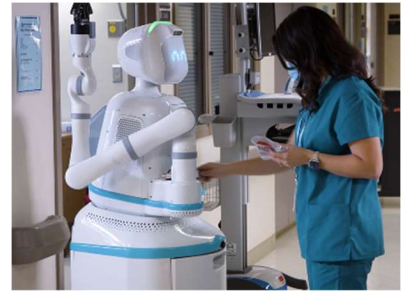

*We envision a future powered by robots that work seamlessly with human teams. We build Artificial Intelligence that enables service robots to collaborate with people and adapt to dynamic human environments. Our robot Moxi currently works in hospitals, executing routine tasks so nurses can focus on what they do best: patient care.*

There are a number of applications for robots in health care. This robot called Moxi assists nurses and staff by performing fetch and delivery tasks in a hospital. It can navigate very well. Their video shows the robot even pushing elevator buttons to be able to go up and down between floors. Meanwhile, the humans can focus on patient care, saving up to 30% of their time. The robot supports the humans; it does not replace them. However, the nurses can't give instructions to Moxi with natural language, because it wouldn't understand them.

## 2.2 Ageless Innovation: Companion Robots

Our country has a growing elderly population, many of whom live alone and have a variety of health issues. Peter has a cousin in her 80s. She is fairly active and has family around, but she lives alone. When he visited her last, he saw her with a couple of these robotic companions. They don't do much, but they do have a little bit of speech capability and can be very entertaining.

A number of companies are producing very simple robots to be companions for Alzheimer's patients and elderly people in general. Ageless Innovation is one company that makes artificial pets that can move, respond to touch, and make comforting sounds. Here are some snippets from their website.

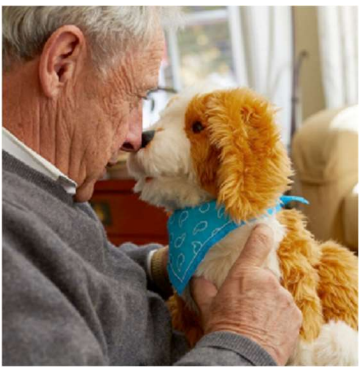

*Companion Pets are designed to bring comfort, companionship and fun to elder loved ones. With realistic fur and pet-like sounds - and sensors that respond to petting and hugs with familiar pet-like actions.*
*Therapy pets have a remarkable effect … in the middle and later stages of Alzheimer's disease. They are calming, soothing and bring the person back to a time in their life when they felt useful and had a sense of purpose; … Alzheimer's pet therapy helps patients escape from a world where someone is always caring for them.*

Imagine a better companion capable of having meaningful conversation and of performing many everyday tasks in the home. They could eliminate the need for 24/7 in-home care and enrich the lives of many people, perhaps any of us one day.

## 2.3 Texas Instruments: Manufacturing Robots

Texas Instruments (TI) is a major manufacturer of electronics that is building a new $11 billion dollar semiconductor fabrication facility in Lehi, Utah. These facilities must be extremely clean inside. Humans work there, but even though they wear a lot of cumbersome clothing, they are still a major hazard since a tiny particle of dust can ruin a chip. To keep things clean, there are also a lot of robots performing various functions.

However, there are some difficulties in implementing the robotics technology. A dean at a technical university within a few miles of this new TI facility recently reported that a manager at TI had asked him to provide 800 trained technicians in the next year to program all the robots. The dean replied that there was no way they could graduate so many trained technicians so fast.

TI also has a whole product line building robots and their components for various industries. They emphasize that these robots are safe and can collaborate with humans.

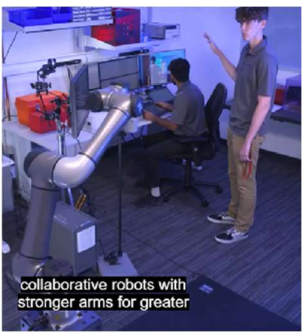

*Bring your robotics designs to life using our products designed for perceptive sensing, precise motor control, real-time communication and AI technologies.*

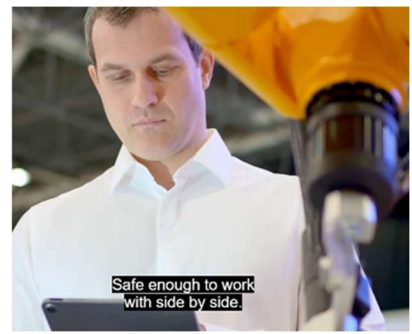

Imagine if the robots in this facility were capable of learning their tasks from experts in semiconductor fabrication that know nothing about programming robots. Then these experts could be robot trainers instead of robot programmers, and the whole system could perform better and be more efficient.

## 2.4 Moley Robotics: Kitchen Robots

Imagine you could have a robot as your personal chef, and it knows how to make all kinds of fancy recipes from world-famous chefs like Andreas Caminada or Tim Anderson. Well, this company has just the robot for you! It knows lots of fancy recipes and can make them to perfection.

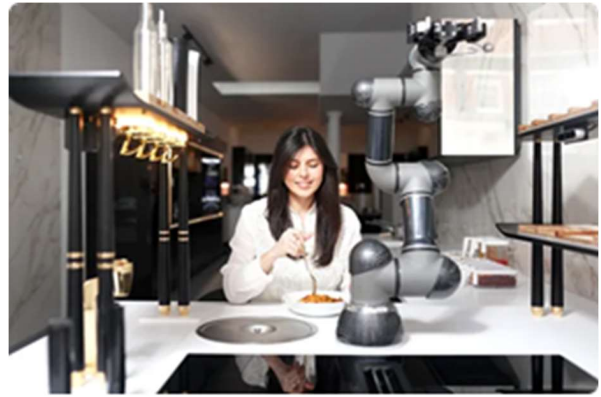

*Moley introduces the Robotic Kitchen, the world's best-selling professional robot, now reimagined for the luxury residential market. Combining state-of-the-art technology with exquisite Made in Italy design, it redefines the standard for culinary living.*

But, what if there's something you don't like about one of those recipes. Wouldn't you like to talk to the robot and tell it how to change things a bit, or even make a whole new recipe of your own creation? That would be much more interesting.

## 2.5 Proven Robotics: Configurable Robot Technologies

This company offers a variety of products and services to help other companies create customized robots for their own applications. They resell at least eight different robots from different manufacturers. They offer services for programming the robots, and even RaaS: Robot as a Service for working with a leased robot on their premises. They provide a software system called PRomMS: Proven Robot Management System that gives a no-code approach to programming a robot using a touch screen.

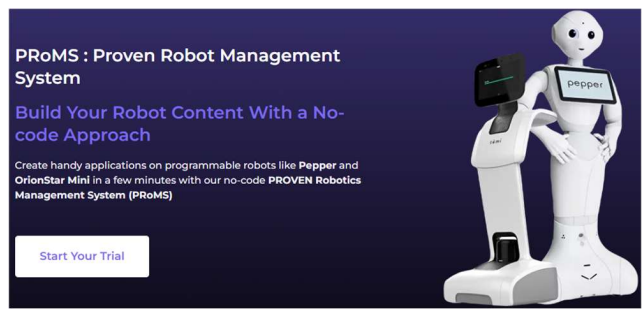

*PROVEN Robotics is bundled with the most competitive robots in the market that have changed the operational landscape, making tasks less redundant and more efficient. We also offer RaaS or Robots-as-a-Service, allowing you to lease our robots for a long or short period.*

This approach provides for a lot of interesting possibilities, but still lacks the ability for the robots to interact dynamically with humans using natural language.

## 2.6 Why language is important and why it is hard

All these commercial robots have the same limitation that we as human collaborators cannot talk to them as we would to a human assistant. In every one of these applications a robust language ability in the robot would raise the collaboration to a much higher level. The question is, how can that be achieved?

Work on Interactive Task Learning (ITL; Gluck & Laird, 2018; Mohan et al., 2012; Lindes et al., 2017) has solved many problems related to building agents that can learn tasks from human instruction using language, but the language abilities of these agents if very limited and would not work well in the real world with human users who are not experts on the limited language the agents have (Ramaraj, 2023).

Why is understanding natural language that addresses objects and actions in the physical world so difficult? (For a review Cohen et al., 2024) We consider three aspects of this problem.

- Grounding object references – Natural language uses referring expressions to identify specific objects for actions. These expressions can be complicated and include things like spatial relations and dialog context. Complex reasoning is required to ground these expressions to their intended objects.
- Relating verbs to tasks and actions – Consider a verb like *cook*. It usually has a direct object to be the thing to be cooked, but how to do the cooking can vary enormously depending on what that object is. The verb could be applied to a single food item, *Cook the potato.*, to a dish requiring multiple ingredients and a complex recipe, *Cook scrambled eggs.*, or to a whole meal for several people, *Cook dinner.* To respond to such a request the agent must have or learn on the spot a lot of knowledge that is very specific to the particular object and the situation.
- Handling free-form language – Traditional ITL agents have useful but very limited language (Lindes et al. 2017; Lindes 2022). A human conversing with another human to teach new tasks will use a wide range of vocabulary, syntax, semantics, and inferred common knowledge that is way beyond the abilities of these agents.

Until recently it seemed that reaching a higher level of language understanding would be very difficult and require a whole new approach to language learning in the agent. Then came LLMs. Now we can envision using them as a powerful assistant to a cognitive agent. In what follows we describe an overall agent architecture that combines a cognitive agent with an LLM and three simple experiments to explore how an LLM might be able to be used to address the three above aspects of the problem.

### 3. A Possible Agent Architecture

Referring again to Figure 1, this section describes in more detail the various parts of the system. It considers how a cognitive agent can receive direction and instruction from a human in natural language, use an LLM to help understand that language and gather elements of general knowledge, gather situational knowledge from its experience, and control and receive perception from a physical robot.

### 3.1 The Human Director

The overall purpose of this AI system is to help a human get tasks done with less effort than would be involved for him or her to do the task themselves. Thus the human is the overall director over the tasks to be performed, as well as the source of understanding their purpose. Although the AI system may provide some knowledge and abilities that the human does not have, the overall collaboration depends on several human abilities the AI system does not have. Using the setting of a private AI chef to help in the kitchen, the human director provides several key abilities.

- Domain expert – Most of all, the human knows what they want, or in other words the purposes of the tasks to be performed.
- Situational knowledge – The human may or may not know a lot about the specific kitchen environment and the details of how to prepare certain meals or food items. If this particular human is lacking some of this knowledge, other means will be needed to gather it.
- Task experience – Often the human will have done similar tasks previously and will have detailed knowledge and heuristics for how to perform them.
- Reasoning – Humans are very good at integrating many different kinds of knowledge and reasoning about it to solve problems or see what would be the best approach to some challenge.
- General knowledge – Humans will usually have a lot of experience and knowledge beyond the particular tasks and domain the robot is intended to work in. Often this knowledge can be applied to solve current challenges.
- Language ability – Humans are extremely capable and flexible in their use of language. They may be able to adapt to deficiencies in the agent's language abilities, but this will be harder for them and take more time.
- Collaboration – Humans usually have experience in collaborating with other humans, using skills of language, of being able to understand the other person's intentions, and of social interactions to make collaborations comfortable and effective. They can apply all these skills to interaction with a robot as well.

## 3.2 The Physical Robot

The embodiment of the system is responsible for perception and action in the world. Both of these are challenging tasks subject to much ongoing research. In the short term, we expect to depend on the AI2-Thor simulation system (Kolve et al., 2022) for these capabilities. It provides simulated rooms with a variety of fixed and movable objects and a simulated robot that can navigate in this world and manipulate objects. The simulation provides high-quality visuals, perceptual inputs to the cognitive agent, and commands to make the robot act in the world.

## 3.3 Situational Knowledge

The cognitive agent contains a short-term memory containing information about the current state of the world and the tasks the agent is performing. It also contains long-term memories which can retain knowledge that the agent learns through its experience and its interactions with the human, the LLM, and the physical world.

This long-term situational knowledge is essential to the agent's ability to learn and perform complex tasks. In our private chef example, this knowledge would include: a map of the physical environment and its fixed objects, knowledge of normal storage locations for both tools and food items, how to clean and/or dispose of all the movable objects, procedures for making a variety of meals and dishes, an inventory of consumable items and how to order replenishments, human preferences for menus and serving protocols, and more. This knowledge must be organized in a way that it can be learned incrementally through experience and used actively while performing tasks.

## 3.4 A Large Language Model

An essential part of the overall system architecture is a large language model (LLM) that the cognitive agent can access. It has three primary purposes in the system.

- Translate language inputs to symbols the agent can understand
- Provide general and common-sense knowledge
- Translate symbols to language outputs

## 3.5 The Cognitive Agent

At the core of this artificial intelligence system is a cognitive agent, in our situation implemented in the Soar cognitive architecture (Laird, 2012; Laird et al., 2012). It provides a number of key capabilities to the overall system.

- Orchestration – Coordinating all the different activities of interacting with the human, controlling the actions of the robot, learning situational knowledge, and making queries of the LLM.
- Long-term knowledge – The Soar architecture provides three forms of long-term memory: a semantic memory for storing facts about the world, an episodic memory for storing its past experiences, and a procedural memory for storing skills that have been either pre-programmed or learned through its experience.
- Learn incrementally – The agent builds its situational knowledge incrementally through interaction with the human, requesting information from the LLM, and reasoning based on its interactions with the physical world.
- Mental models – Soar's memories and reasoning capability give it the ability to build internal models of the physical world, of actions and their effects, of human interactions and expectations, of tasks to be performed, etc.
- Reasoning – The Soar agent has reasoning capabilities that allow it to evaluate inputs, plan sequences of actions to achieve its goals, and many other things.
- Verifying the LLM – These reasoning capabilities combined with its long-term situational knowledge give the agent the ability to verify responses to its LLM queries to see if they are consistent with its knowledge of the current situation. It can also ask the human collaborator to confirm a course of action or new fact retrieved from and LLM before acting on it.
- Integration – This cognitive agent is the central element that integrates the physical capabilities of the robot, the knowledge and translation resources of the LLM, and its own situational knowledge and reasoning abilities in a way to be able to collaborate effectively with a human, performing tasks as instructed and learning the knowledge to become more and more autonomous as it goes.

The first author is part of a research group that at the Center for Integrated Cognition.[1] This group is doing research, so far unpublished, on an intelligent robotic system similar to what is described here. It is based on the Soar cognitive architecture and it is integrated with the AI2-Thor simulation system. So far this project has done little work on integrating the LLM with this particular agent, but we have done several other projects related to a cognitive agent gathering knowledge from an LLM (Kirk et al., 2024[2], 2023). James Kirk (Kirk, 2024) of that group did some early preliminary experiments on using an LLM to ground language, and these have been an inspiration for the experiments reported here.

## 4. Challenge #1: Identifying Objects from Referring Expressions

As a human gives instructions to our agent on how to do things and provides additional situational knowledge, a large part of the natural language used consists of referring expressions (REs) intended to identify specific objects that are immediately visible or objects that the agent may have to seek for. Language coming from an LLM in response to queries for general or common-sense knowledge will also have many referring expressions. Grounding these referring expressions to specific objects in the real world is a significant challenge that must be met (Lindes & Skiker, 2025).

### 4.1 Seven ways to identify an object

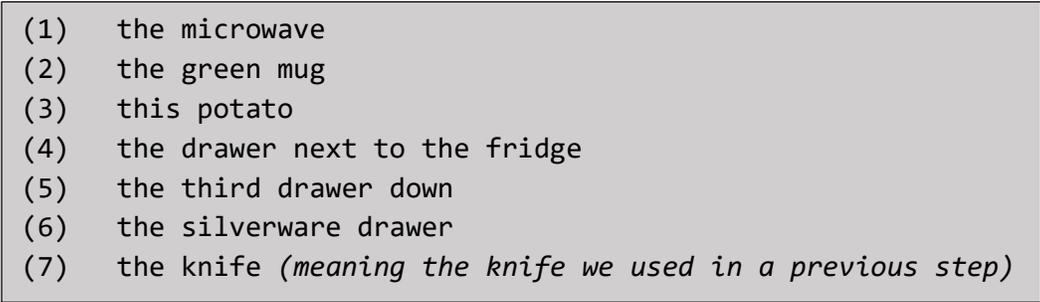

```
(1)   the microwave
(2)   the green mug
(3)   this potato
(4)   the drawer next to the fridge
(5)   the third drawer down
(6)   the silverware drawer
(7)   the knife (meaning the knife we used in a previous step)
```

*Figure 2: Seven forms of referring expressions*

Figure 2 gives examples of seven different classes of referring expressions. The most general is (1), where a noun is used to refer to a category of objects. (2) illustrates adding adjectives that select according to physical properties of a particular object. A human may use pointing gestures or similar things to refer to an object deictically, as in (3). Prepositional phrases, as in (4), are often used to specify an object using a spatial relation to some reference object. Other kinds of relative spatial relations can be expressed with adjectives, as in (5). Often objects are identified by their intended function or use, as in (6). In the course of a dialog, a reference like (7) can be used to identify an object that is currently salient in the recent dialog. Our system must be able to correctly ground object references based on all seven of these general patterns.

## 4.2 Analysis of Referring Expressions

Humans often use single category names, such as *the microwave* to refer to objects in the environment. In simple scenes, where the number of objects is limited and their categories are distinct, such references are generally sufficient for identification. Mapping these category names to object categories in the agent's world model can allow the robot to resolve such expressions reliably. Preliminary experiments (Kirk, 2024) suggest that modern LLMs can successfully ground language references to internal object identifiers based on category names, when provided with a structured list of objects.

However, category names alone are often insufficient in more complex environments, like kitchens or bedrooms, where many objects share the same category (e.g., multiple drawers or cups), making it impossible to disambiguate the target object using categorical information alone. In these cases, humans naturally use other types of referring expressions that combine multiple features to help distinguish one object from another (see Figure 2). The ALFRED dataset (Shridhar et al., 2020) shows how humans combine perceptual, functional, and spatial attributes for describing objects using language. Table 5 shows some examples of how different humans adapt their REs to the context of the environment and choose the appropriate distinguishing features.

Perceptual REs such as *the red cup* or *the square box* is another kind of expressions humans use often to refer to objects in the physical world. They are usually expressed using an adjective followed by a noun. Grounding these REs requires information about the perceptual properties of objects such as color or shape abstracted from the Robot's camera. The cognitive agent must be able to associate descriptors in language or adjectives such as *red* and *square* with the perceptual features in its internal representation of the physical model. Table 5 shows that humans often use such REs to refer to objects in different settings. However, perceptual REs can be sometimes ambiguous and complex to ground as they may refer to several objects. Spatial REs such as *the cup on the left* or *the cabinet next to the stove* have received considerable attention (Mohan et al., 2025; Mi et al., 2020; Tellex et al., 2011). These expressions require the agent to possess explicit knowledge about spatial relationships between objects. A scene graph, a graph-based structure composed of nodes and edges, has been employed to resolve spatial REs (Mi et al., 2020). Such graphs enable the cognitive agent to reason about spatial REs by explicitly encoding spatial relationships, where nodes represent objects in the environment and edges denote relations between them (e.g., left-of, above, next-to). A problem with this approach is that in a scene with hundreds of objects and with dozens or more possible linguistic terms for relations the graph becomes intractable.

Tellex et al. (2011) introduced a probabilistic grounding graph that allowed robots to map spatial prepositions like *on, next to*, or *under* to actual locations in their environment. Shridhar and Hsu (2018), however, adopted a different approach: grounding by generation. Their system, INGRESS, is a two stages neural-network model that in the first stage generates visual descriptions of objects, matches them against the input expression, and identifies a set of candidate objects. In the second stage it examines all pairwise spatial relations between the candidates to infer the most likely referred object.

In our own work, we apply a related idea by constructing neighbor graphs from a world model. These graphs capture spatial relations between objects and allow the cognitive agent to perform symbolic reasoning to resolve REs such as *the cabinet to the left of the microwave* or *the third drawer down next to the stove*. One major problem that can arise when analyzing spatial REs is prepositional ambiguity. The instruction *Put the apple in the towel in the box* is ambiguous because it is unclear whether the phrase *"in the towel"* specifies the apple's current location or describes the goal location of the verb put. In that case, it is necessary to integrate the syntactic parsing with contextual and world knowledge to infer the most likely interpretation (Lindes et al., 2017). Functional REs, such as *the drawer with the knives* or *the cabinet where the cups are stored* are more complex. They require situational knowledge: knowledge of how the environment is organized or how objects are typically used. Unlike perceptual and spatial features, this information is not directly observable and must be learned through human interaction or some other means. For instance, resolving *get a knife from the drawer* assumes knowledge about which drawer contains the knives. Such knowledge is not typically embedded in an LLM. A cognitive agent must then acquire it through human interaction to reason over functional REs.

Deictic references such as *put it over there* or *use this knife* require integrating additional perceptual input about human pointing or shared gaze.

Dialog references like *do the same as before* are another important class of references that humans use to talk about previous actions or objects. For example, if an agent has previously sliced a potato with a knife and now needs to use the knife again, it must infer that *the knife* refers to the same object it used earlier. This kind of RE depends on dialogue history. In the Soar cognitive architecture, episodic memory contains such knowledge to track object usages. Modern LLMs like ChatGPT (OpenAI, 2023) also offer a form of dialog memory through their context windows, which may allow them to resolve such references. We are currently exploring the effectiveness of both approaches.

One promising  approach is to use the capabilities of large language models (LLMs) to help a cognitive agent resolve complex and ambiguous REs. These models are trained on vast corpora of human texts and encode a large amount of commonsense knowledge. Given sufficient context, LLMs can generate or interpret REs by identifying which distinguishing features are most appropriate to refer to a unique object in this context. Despite these capabilities, no unified approach has yet emerged that can handle all types of REs across all environments. The challenge remains to combine perceptual, spatial, functional, episodic, and commonsense knowledge in an integrated solution.

## 4.3 Initial Experiments

To address this gap, we propose a combined approach that integrates the reasoning abilities of a cognitive agent (in the Soar architecture) with the language capabilities of an LLM. In our approach, the LLM interprets the input referring expression and either 1) performs the full resolution itself when supplied with additional structured knowledge or 2) produces a formal representation suitable for symbolic resolution by the cognitive agent. We conducted several variations of an experiment to evaluate the first approach.

The context for these experiments is a kitchen scene provided by AI2-Thor. Figure 3 shows two views of this kitchen with labels with arbitrary numbers applied to the cabinets and drawers[3]. Each of these object categories has a number of instances.

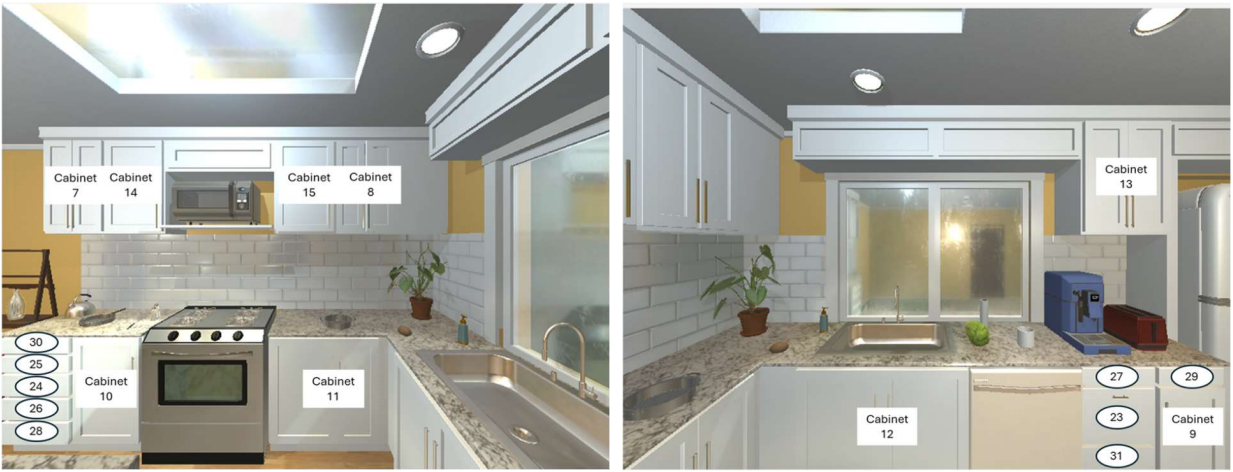

a) Objects with labels looking South     b) Objects with labels looking West

*Figure 3: Two views of a kitchen*

Given this context, we wrote a set of referring expressions to refer to cabinets and drawers to see how well an LLM could resolve the ambiguities arising from multiple objects of the same category.

*Table 1: Referring expression tests*

| Number | Text | Challenges |
|--------|------|------------|
| 1 | the high cabinet to the left of the microwave. | high, left |
| 2 | The cabinet second to the right of the microwave. | second, right |
| 3 | the top drawer next to the stove. | top, next to |
| 4 | the drawer next to the fridge. | next to |
| 5 | the middle drawer next to the dishwasher. | middle, next to |
| 6 | the low cabinet next to the fridge. | low, next to |
| 7 | the cabinet below the sink. | below |
| 8 | the fourth drawer to the left of the stove. | fourth, left |
| 9 | the bottom drawer next to the dishwasher. | bottom, next to |
| 10 | The second cabinet to the left of the microwave. | second, left |

---

[3] The labels for the drawers are abbreviated in the figure. The actual labels are like "Drawer23".

These test expressions have challenges in resolving spatial relations (4) and spatial properties (5). To test whether an LLM could correctly ground these expressions given appropriate context, a set of eight tests, labeled A to H, were run with ChatGPT as it was on 1 August 2025. Each test had a different variation of a common prompt. The general structure of the prompt is given in Figure 4.

```
I am a robot trying to talk with a human, and I need help understanding the human's
language. Specifically, I need help ground referring expressions to objects I can
see in my world.

<optional category list>

<neighbor graph>

If I give you a referring expression, I want you to return the object id of the
correct object.
Example:
Input: 'the countertop left to the microwave'
Output: CounterTop17

Here are the referring expressions I would like you to ground for me.
For each give me a single line of text with the referring expression followed by the
object id it refers to.
If more than one matches, give all the matches.

Input referring expressions:

<the ten referring expression tests>

Please give me your answers:
```

*Figure 4: Prompt for object grounding experiment*

*Experiment Variations*

The experiments provided ChatGPT with different combinations of two forms of the category list and three forms of the neighbor graph, some with additional explanatory text. The list of referring expressions was always the same, as shown in Table 1. Table 2 shows how these variations were distributed over the eight tests, and below we explain the details.

*Table 2: Prompt variations for eight tests*

| Test | Category | Neighbors | Explanation |
|------|----------|-----------|-------------|
| A | IDs | language | |
| B | none | language | |
| C | IDs | x, y, z | |
| D | none | x, y, z | |
| E | IDs | x, y, z | In this neighbor graph +x is to |
| F | none | x, y, z | the left and +y is to the right. |
| G | Objects | N,S,E,W | (N:North S:South W:West E:East U:Up D:Down) |
| H | Objects | N,S,E,W | (N:North S:South W:West E:East U:Up D:Down) spatial relations are viewpoint dependent *(Neighbors given from two different labeled viewpoints)* |

*Category Lists*

The primary tool for identifying objects from language is the category, usually expressed as a noun. This is item (1) in the list in Figure 2. A list of categories for all the objects was given in all the tests except B, D, and F. It had two possible forms. For tests A, C, and E the category IDs looked like these examples:

```
Microwave43 : category Microwave,
Cabinet7 : category Cabinet ,
...
Stove78 : category Stove ,
Drawert23 : category Drawer ,
...
```

For texts G and H this list was changed to replace the object IDs on the left with unique identifiers that had no useful meaning, as in these examples:

```
Object7: category Cabinet
Object43: category Microwave
...
Object30: category Drawer
Object78: category Stove
...
```

This is intended to make the job harder for the LLM since it can't infer the object category from its identifier. An additional reasoning step is required.

The neighbor graph is the key to being able to ground objects based on spatial relations and spatial modifiers, items (4) and (5) in the list in Figure 2. Over the different experiment variations, the neighbor graphs were given in three different forms: with language, with x, y, z coordinates, or with cardinal directions. Examples of the language variation looked like this:

```
Microwave43 (left= Cabinet14, right= Cabinet15, below= Stove78)
Cabinet15 (right= Cabinet8, left= Microwave43, below= CounterTop19a)
Cabinet8 (left= Cabinet15, below= Counterabove19a)
Cabinet14 (right= Microwave43, left= Cabinet7, below= Counterabove17)
Cabinet7 (below= Counterabove17, right= Cabinet14)
...
```

Examples of the x, y, z variation looked like this:

```
Microwave43 (+x= Cabinet14, -x= Cabinet15, -z= Stove78)
Cabinet15 (-x= Cabinet8, +x= Microwave43, -z= CounterTop19a)
Cabinet8 (+x= Cabinet15, -z= Counter+z19a)
Cabinet14 (-x= Microwave43, +x= Cabinet7, -z= Counter+z17)
Cabinet7 (-z= Counter+z17, -x= Cabinet14)
...
```

Examples of the cardinal direction variation looked like this,  in JSON format:

```
{
  "Object7":{ "W":["Object14"], "D":["Object17"] },
  "Object8":{ "E":["Object15"], "D":["Object19a"] },
  "Object9":{ "N":["Object36"], "S":["Object23","Object31"],
    "U":["Object29"] },
  "Object10":{
    "E":["Object30","Object25","Object24","Object26","Object28"],
    "W":["Object78"], "U":["Object17"] },
  ...
}
```

## 4.4 Results and Analysis

Table 3 shows the results for each of eight variations of the experiment. Yellow cells are incorrect answers. Now we attempt to understand why different kinds of referring expressions are difficult for the LLM to ground depending on the prompt variations.

*Table 3: Groundings given by ChatGPT*

| RE | Correct | A | B | C | D | E | F | G | H |
|----|---------|----|----|----|----|----|----|----|----|
| 1 | 14 | 14 | 14 | 15 | 15 | 14 | 15 | 14 | 14 |
| 2 | 8 | 8 | 8 | 7 | 7 | 7 | 7 | 8 | 8 |
| 3 | 30 | 30 | 30 | 30 | 30 | 30 | 30 | 30 | 30 |
| 4 | 29 | 29 | 29 | 29 | 29 | 29 | 29 | 29 | 29 |
| 5 | 23 | 27 | 23 | 27 | 23 | 27 | 27 | 23 | 23 |
| 6 | 9 | 9 | 13 | 9 | 9 | 9 | 9 | 13 | 13 |
| 7 | 12 | 12 | 12 | 12 | 12 | 12 | 12 | 12 | 12 |
| 8 | 26 | 24 | 24 | 26 | 26 | 26 | 26 | 28 | 24 |
| 9 | 31 | 31 | 31 | 31 | 31 | 31 | 31 | 31 | 27 |
| 10 | 7 | 7 | 8 | 8 | 8 | 8 | 8 | 14 | 7 |

In this situation the LLM is only performing part of the grounding process for spatial relations. The LLM does not do geometric calculations based on position data. Instead, the cognitive agent does an initial analysis to produce the list of categories and the neighbor graph. Then it gives this information to the LLM so that the LLM can apply its knowledge of language and a small amount of reasoning to ground referring expressions involving spatial relations and properties. Although the LLMs language capabilities are great, its reasoning abilities are limited. Our eight experiments can give some idea of where those reasoning limitations are.

It would be desirable to have a system of measuring the difficulty of processing a specific referring expression relative to the kind of information provided. Using a somewhat arbitrary and intuitive scoring system, we assigned points to the number of reasoning steps required to ground a particular referring expression, based on the properties and relations shown as challenges in Table 1. Then we assigned additional points based on the degree of difficulty determined by the information provided in each experiment variation. Although these numbers are very arbitrary, they can give some idea of the amount of reasoning the LLM needs to perform. Table 4 shows the results of applying this point system to each referring expression for each variation.

*Table 4: Difficulty scores*

| RE | Experiment Points | A | B | C | D | E | F | G | H |
|---|---|---|---|---|---|---|---|---|---|
|  |  | 2 | 1 | 4 | 3 | 3 | 2 | 5 | 5 |
| 1 | 6 | 8 | 7 | 10 | 9 | 9 | 8 | 11 | 11 |
| 2 | 7 | 9 | 8 | 11 | 10 | 10 | 9 | 12 | 12 |
| 3 | 6 | 8 | 7 | 10 | 9 | 9 | 8 | 11 | 11 |
| 4 | 5 | 7 | 6 | 9 | 8 | 8 | 7 | 10 | 10 |
| 5 | 9 | 11 | 10 | 13 | 12 | 12 | 11 | 14 | 14 |
| 6 | 7 | 9 | 8 | 11 | 10 | 10 | 9 | 12 | 12 |
| 7 | 4 | 6 | 5 | 8 | 7 | 7 | 6 | 9 | 9 |
| 8 | 9 | 11 | 10 | 13 | 12 | 12 | 11 | 14 | 14 |
| 9 | 7 | 9 | 8 | 11 | 10 | 10 | 9 | 12 | 12 |
| 10 | 7 | 9 | 8 | 11 | 10 | 10 | 9 | 12 | 12 |

In this table the green shaded cells are the scores given to the referring expressions, the blue shaded ones are the additional points given to each experiment variation, and cells inside the matrix show the total score for each RE/experiment combination. Yellow shading is on the cells where the LLM got the wrong answer as shown in Table 3.

Of course, this scoring system is very rough and ignores considerations that an LLM uses probabilities and does not always give the same result for the same question. Nevertheless, Table 4 suggests that when the scores we assign get to around 10 to 12, the LLM's reasoning becomes inadequate, at least for the version of the LLM used for these tests.

These results suggest that the LLM can use structural knowledge in the form of category and neighbor information to resolve referring expressions in natural language that contain spatial expressions. Using this capability, it could effectively collaborate with a cognitive agent that can provide it with the necessary symbolic knowledge. The results also show that this capability of an LLM tends to break down when the number of reasoning steps becomes too great.

## 4.5 Ways to Improve

These experiments illustrate the fact that grounding referring expressions involving spatial relations and properties really involves three parts:

1) Performing mathematical operations on position data to create symbolic representations of spatial relations.
2) Converting a wide range of natural language expressions into logical operations that can be used to ground these expressions to specific objects using these symbolic representations.
3) Performing the reasoning defined by these logical operations to compute the actual groundings of each expression.

In the experiments documented here, we assumed the cognitive agent could provide the symbolic data for 1) and that the LLM could do 2) and 3). Our results show that the LLM is not that good at 3), and we already know the cognitive agent is weak in doing 2). Another possible approach would be to ask a different kind of question to the LLM. Instead of asking it to do both 2) and 3), suppose we were to ask it to just convert each referring expression into a set of logical operations and then have the cognitive agent do the reasoning of step 3).

The CIC research group has done some yet unpublished experiments in a different domain to study this kind of approach, and the results look promising. If we can define adequately the set of logical operations needed, and give the LLM the necessary prompts to have it produce  the operations to perform for each language expression, this approach should work. It makes the best use of both the LLM and the cognitive agent. We expect pursuing this approach can produce more reliable and more correct results.

## 5. Challenge #2: Performing Complex Tasks

The tasks involved in managing a kitchen or another complex environment typically involve a lot of different kinds of objects and many different actions that can be performed on them. Most action verbs will define multiple different variations of a task depending on the type of object or objects involved. A verb like *discard* for example could mean putting something in the garbage can, the recycling bin, or the sink depending on what the object is. A verb like *cook* has a much wider range of possible task variations for a wide range of possible direct objects of the verb. This section considers some possible ways to deal with these variations.

### 5.1 Examples

Consider the situation of a robotic household assistant. Its owners would like it to take charge of the kitchen and many other routine household tasks. When first purchased, this agent has only some general household knowledge and skills. The family members gradually teach it things about the routines of this particular household, where to find and store various kinds of objects, and how to perform a variety of tasks according to their preferences. Over time the agent becomes a valuable helper. We would expect it to eventually know how to perform all the following tasks without further instruction:

```
Cook the potato.
Cook my breakfast.
Cook dinner.
Tidy the kitchen.
Clean the bathroom.
Make the beds.
Replenish supplies weekly.
```

These are all task commands that consist of an action verb and an argument. Other task commands might have a second or even third argument. However, the details of how to perform the task depend very much on what kinds of objects the arguments refer to, and other aspects of the environment and immediate situation. Therefore, the relationship between a particular action verb, like *cook*, and a detailed action plan is very complex and involves a lot of additional knowledge.

The cognitive agent will need to learn all the varieties of these tasks, as well as the background situational knowledge needed to perform them. Where can it acquire all this knowledge? Our vision is that it can learn the necessary knowledge incrementally from a large number of experiences over time. We expect that as the agent is attempting to collaborate with the human it will reach impasses where it does not know what to do next. Then it will reach out to the either the human or the LLM to ask for help. When it has successfully resolved that particular impasse, it will learn long-term knowledge that it can use the next time it encounters a similar situation. As the process continues, the agent's competence grows.

## 5.2 Analysis

Understanding verbs in context remains an open challenge for cognitive agents as they require a vast amount of knowledge and must be translated into non-linguistic action primitives that can directly be executed by the physical robot. In this chapter, we explore both problems, present prior works and suggest some potential solutions.

*Knowledge Extraction and Reasoning*

Understanding  action verbs requires a vast amount of knowledge, some of which must be learned incrementally through human interaction (Kirk et al 2023b; Chen et al. 2020), and some retrieved from external knowledge databases (Wray & Kirk, 2021; Kirk et al 2023a; Gao et al.2024). Consider the instruction *cook the potato*: to act on it, the agent must know what cooking entails, what tools are involved, where to find potatoes, and possibly user preferences. This type of reasoning is non-trivial and deeply context dependent.

Studies such as Kirk et al. (2023a, 2024) have investigated how LLMs  can be used to extract relevant procedural knowledge for novel tasks. Yet, the complexity and variability of LLM-generated responses pose a serious challenge for a cognitive agent. With the careful design of prompts, an LLM can help by providing knowledge that is understandable to the cognitive agent for a given context (Kirk et al., 2022; Yu et al., 2023).

Interpreting everyday human instructions such as *I'm hungry, what's for breakfast?* or *Cook me an egg* requires an agent to understand not just the language, but the surrounding situational knowledge in the current environment: recipes, available ingredients, storage locations, tools, and social context. LLMs can't ground that sort of language because it is very context dependent. A cognitive agent can accumulate and learn such knowledge from humans through interactive learning.

*Translating Verbs to Action Primitives*

Action verbs must be decomposed into low-level motor commands or non-linguistic action primitives executable by the robot. This translation process is highly non-trivial because most verbs abstract over multiple procedural steps. For instance, *"clean up"* may involve identifying a mess, retrieving cleaning tools, and wiping a surface, and much more knowledge.

Early efforts to understand verbs in context can be traced back to Winograd's SHRDLU system (1972)**,** that uses natural language instructions such as *put the red block on the green cube* to manipulate blocks in a simulated environment. In this system, verbs such as *put* are associated with procedures of the steps needed to perform that action. However, this system is limited to understand a small set of predefined verbs that was explicitly programmed for a blocks world.

To process a large set of verbs, Lindes & Laird (2017) used Embodied Construction Grammar (ECG; Bryant 2008) to represent their meanings. In this approach, each verb is linked to a semantic schema that defines its event structure, the roles of its participants, and any constraints on the action. These schemas are connected to a world model so the agent can interpret the verb in context, check whether the action is possible, and generate an executable plan. This integration allows the system to disambiguate verb meanings using both linguistic information and knowledge of the environment (Chen et al., 2011).

Another approach to this problem is interactive task learning (ITL; Gluck & Laird, 2018) where humans guide robots through new tasks step-by-step (Lindes et al., 2017; She et al., 2014). In ITL, humans engage in a dialogue with a cognitive agent to help it learn new verbs (Cantrell et al., 2011; Chai et al., 2018). Research by She et al. (2014) and Lindes & Laird (2017) shows how teaching-by-demonstration can help robots learn new verbs such as *"tidy up"* by breaking them into a series of executable sub-actions. Another way approach this is to invite humans via Amazon Mechanical Turk to describe video demonstrations of everyday tasks such as pick and place and provide the goal instruction composed of one verb with the detailed description of that goal involving several actions verbs as in ALFRED (Shridhar et al., 2020).

Through prompt engineering, a large language model (LLM) can decompose a high-level instruction into a sequence of low-level actions. For example, Ahn et al. (2022) introduced a system called *SayCan*, which can interpret instructions like *I spilled something, can you help?* and break them down into specific low-level actions, such as: 1) find a sponge, 2) pick up the sponge, 3) bring it to you, and 4) indicate completion (Ahn et al., 2022). Lawley & MacClellan (2024) report on an agentic LLM system called VAL with similar goals to ours. They highlight the necessity to break up a complex request to an LLM into small pieces they call "subroutines." But VAL lacks the abilities of a full cognitive agent such as reasoning, memory and language understanding.

## 5.3 Experiments

Gathering incrementally all the knowledge needed to perform tasks like those given in the examples above will be a long process. Often it will be possible to get additional general or common-sense knowledge from the LLM. Here we present a simple initial experiment to gather knowledge on how to *store* objects.

*Store* is a verb that usually has a direct object of the thing to be stored and an implicit object of the location to store it. That implicit location will vary a lot depending on what the direct object is. In this experiment we give ChatGPT (the free version on 8/8/2025) a prompt containing several types of information: 1) a list of objects with their categories; 2) a description of the problem to decide where each object should be stored; 3) a list of general types of objects; and 4) a list of referring expressions to describe objects that we want to know where to store. The list of objects includes both objects to be stored and possible storage locations. Figure 5 shows the entire prompt with the object list abbreviated. Figure 6 shows the response from ChatGPT.

```
I am a robot trying to talk with a human, and I need help understanding the human's
language. Specifically, the human has asked me to store things, and I don't know
what store means. Here are the objects I can see:
World State:
object1: category apple
object2: category book
object3: category bottle
object4: category bowl
...
object82: category vase
object83: category window
object84: category wine-bottle

Here are some referring expression for things the human wants me to store. For each
give me a single line of text with the referring expression followed by the object
id of the object to be stored and the object id of the location to store it. If more
than one matches, give all the matches.
There are several types of objects that I want to be able to store:
Type 1: Perishable food items.
Type 2: Cooking tools.
Type 3: Eating utensils.
Type 4: Dishes
Type 5: Pots and pans
Type 6: Cleaning tools
Input referring expressions:
the apple
the spatula
a sponge for washing dishes
a bottle of wine
a vase
a fork
potatoes
a butter knife
a soup spoon
a roll of paper towels
my favorite  plant
something to slice the bread

Please tell me where would be a good place to store each of these objects:
```

*Figure 5: Prompt for asking LLM where to store objects*

```
Here is the interpretation of each referring expression with the format:
[Referring Expression] → [Object ID to Store] → [Object ID(s) of Suggested Storage
Location(s)]
________________________________________
Type 1: Perishable food items
(These are typically stored in the fridge or possibly cabinets for non-refrigerated
items.)
•     the apple → object1 → object40 (fridge)
•     a bottle of wine → object84 → object40 (fridge)
•     potatoes → object54 → object55 → object40 (fridge)
•     my favorite plant → object42 → object58, object59, object60 (shelves)
________________________________________
Type 2: Cooking tools
(These are usually stored in drawers or sometimes cabinets.)
•     the spatula → object65 → object27, object28, object29, object30, object31,
object32, object33, object34, object35 (drawers)
•     something to slice the bread → object44 (knife) → object27, object28, object29,
object30, object31, object32, object33, object34, object35 (drawers)
________________________________________
Type 3: Eating utensils
(These go in drawers or utensil organizers.)
•     a fork → object39 → object27–object35 (drawers)
•     a butter knife → object6 → object27–object35 (drawers)
•     a soup spoon → object66 → object27–object35 (drawers)
________________________________________
Type 4: Dishes
(Stored in cabinets or shelves.)
•     a vase → object81, object82 → object58, object59, object60 (shelves) or
object7–object15 (cabinets)
________________________________________
Type 5: Pots and pans
(Store in cabinets, lower shelves, or pot drawers — none explicitly requested.)
(Not applicable in current referring expressions.)
________________________________________
Type 6: Cleaning tools
(Stored under sink, in cabinets, or in designated cleaning storage.)
•     a sponge for washing dishes → object25 → object7–object15 (cabinets)
•     a roll of paper towels → object50 → object7–object15 (cabinets)
________________________________________
Let me know if you want to simulate storing these or need location preferences based
on common kitchen layouts.
```

*Figure 6: ChatGPT response on where to store objects*

The response is much too verbose for what we want our agent to deal with, but
previous experience has shown that it is easy to modify the prompt to get a much more
concise response. Beyond that, the response shows that ChatGPT is providing some
general knowledge that could be useful to our agent.

In analyzing the response in Figure 6 we see several things. ChatGPT is putting all the requested objects into the type classes we defined based only on the language we gave it and its internal general knowledge. Once this has been done, it also uses its general knowledge to suggest the proper storage location for each requested object. All of this knowledge could be very useful for our agent.

However, we see problems in this response as well. Potatoes are not really perishable enough to go in the fridge, and a house plant is not a food item. These issues are a function of the limited set of types we gave in the prompt. Many things are said to be stored in a drawer or a cabinet without specifying which one. Again, we could blame this on our prompt not giving any information on what storage locations are appropriate for what type of objects. The LLM gave us useful knowledge, but it is not complete and not completely accurate.

How can our agent deal with the limitations like these of LLM responses? The answer is in the integrated system shown in Figure 1. As the agent goes about trying to store things based on the knowledge it got from the LLM, it has ways it can confirm and/or correct this knowledge. When storing a particular object, the cognitive agent can first examine its episodic memory of past experience to see where that object has been found before. It can also ask the human to confirm the storage location before actually putting the object there, and correct the storage location if needed. Another option is to go ahead and put the option where the LLM specified, after which the human can interrupt the agent to correct it if that was not correct. The key to solving problems of this sort is to combine the strengths of the human and the LLM using the reasoning and orchestration abilities of the cognitive agent.

## 6. Challenge #3: Understanding Free-form Language

Humans often use free-form language to give instructions. Unlike formal language, free-form language does not necessarily follow strict grammatical rules or fixed semantic structures. In addition, humans often use a large vocabulary and a wide range of syntactic forms and semantic inferences. This type of language is common in natural conversations and reflects how people typically speak in everyday life. While good language understanding would be very useful for human-robot collaboration, understanding free-form language remains a significant challenge for cognitive agents.

Previous work (Lindes et al. 2017; Lindes 2022) has shown that a Soar cognitive agent can be programmed to understand a set of sentence structures using a limited vocabulary and a finite amount of syntactic and semantic knowledge. By composing these elements in different ways, the number of possible sentences is very large. Nevertheless, everyday human conversation has many forms such a system cannot begin to understand. Suppose that we could use an LLM to translate natural human English into a restricted English the agent could understand. That would be a big step forward. Here we take an initial look at how that might work.

## 6.1 Examples

*ITL Examples*

The CIC research group is beginning to look at using language to instruct our latest robotic agent. Figure 7 shows some examples of task instructions we are using for testing. There are several things that the agent will find difficult to understand:

- Complex task names, like *make, store, clean up, explore, serve*. These will require a substantial amount of interactive instruction or learning from an LLM to understand how they apply to different kinds of objects and what action steps are required to perform them.
- Complex sentence structures with multiple clauses
- Conjunctions of objects and commands
- Disjunctions of objects and commands
- Quantifiers like *some, a few*, or *all*

```
Store Potato55 and the apple in the fridge.
Make toast.
Get an apple and put it in the fridge.
Open the microwave, then get the potato, and then close the microwave.
Put the apple in the fridge or on CounterTop22.
Put the apple in the fridge or on CounterTop22 and turn on the faucet
Clean up the counters.
Explore the receptacles.
Close the doors.
Put the apple in the microwave, then close the door. Put the potato and then the
apple in the fridge.
Put the apple in the microwave, then the potato in the fridge. Swap their locations,
and then cook the potato, and then return them to their original locations.
Put the apple in the microwave, then the potato in the fridge. Swap their locations,
and then return them to their original locations.
Put the apple in the fridge. Close the door, then put the apple on one countertop
and the egg on another.
Put the plate on the countertop, then put the egg on the plate and cook it in the
microwave. Move the egg and plate to the countertop.
Make toast then cook an egg and make coffee. Serve them all.
```

*Figure 7: Examples of human task descriptions*

*ALFRED Examples*

To test and refine our system, the human annotations found in the ALFRED dataset (Shridhar et al., 2020) are a valuable source of training and evaluation data. These annotations offer real-world examples of the kind of free-form expressions we aim to handle, including varied references to objects and different verb usages for the same actions.

The ALFRED dataset has over 8,000 "demonstrations" of a robot in AI2-Thor performing one of seven types of tasks. Amazon turk "annotators" observed videos of the agent performing each task, and were asked to write both a detailed description of the actions the agent took and a single sentence describing the goal of the task. Each demonstration has such annotations made by three different human subjects.

Figure 8 shows three different annotations made by three different humans after all had seen the same video of a simple task. These texts illustrate well that there is a great deal of variation in how humans express a description of the same experience. Between the three annotations one finds a variety of vocabulary, many different sentence structures, and many different referring expressions that refer to the same object. Even within the same paragraph the same person used different ways of referring to the same object, as shown in Table 5. Clearly the new system we are proposing needs a much more robust way of understanding natural language than what has been used in our traditional ITL systems (Lindes et al., 2017).

| | | |
|---|---|---|
| **Goal:** Place a slice of cooked potato onto the counter. | **Goal:** Put a cooked potato slice on the counter. | **Goal:** Put a piece of cooked potato on the counter. |
| **Description:** Turn right, move to the table. Pick up the knife from the table. Slice the potato on the table. Turn left, move to the counter left of the bread. Put the knife on the counter near the soap container. Turn left, move to the table. Pick up a slice of potato from the table. Turn left, move to the counter in front of the stove. Put the potato slice into the microwave, cook it, pick it back up. Turn right, move to the counter left of the bread. Put the cooked potato slice on the counter. | **Description:** Turn right, turn right, walk past the sink, turn left to face round table with tablecloth. Pick up the yellow-handled knife from the table. Cut a slice in the potato on the table. Turn left, turn left, turn right at counter, cross room, turn left at refrigerator to face counter. Put knife down on the table. Turn left, walk past sink, turn left to face round table. Pick the potato slice up from the table. Turn left, make right around corner of counter, turn left to face stove and microwave. Put potato in microwave, cook it, take it out of microwave. Turn right, cross room, turn left at counter with blue plate on it. Put potato on the counter in front of the blue plate. | **Description:** Turn right and cross the room, then turn left and go to face the gray table. Pick up the knife from in between the lettuce and the apple. Use the knife to slice the potato that's on the gray table. Bring the knife with you and go face the kitchen counter with the loaf of bread. Put the knife down in front of the soap dispenser on the counter. Go back over to the gray table. Pick up a slice of the cut potato from the table. Bring the potato with you and go over to the stove, then look up at the microwave. Cook the potato slice in the microwave, then take it out again. Bring the potato slice over to the countertop with the loaf of bread and the knife you used to cut it. Put the potato slice down in front of the blue plate. |

*Figure 8: Three human annotations for the same task*

|  | Human1 | Human2 | Human3 |
|---|---|---|---|
| Table | Round table with tablecloth Round table | The table | The gray table |
| Knife | Yellow handled knife | The knife | The knife from in between the lettuce and the apple The knife The knife used to cut the potato |
| Potato | A slice in the potato | The potato slice The potato A slice of potato The cooked potato slice | The potato on the gray table A slice of the cut potato The potato slice |
| Counter | Counter counter with blue plate on it | the counter near the soap container the counter in front of the stove the counter left of the bread | The kitchen counter with the loaf of bread The counter top with the loaf of bread. |

## 6.2 Analysis

The use of free-form language in human-robot collaboration is very promising. On one hand, it allows humans to communicate naturally, using the kind of language they are already comfortable with. On the other hand, it enables people who are not experts in robotics or programming to interact with robots in meaningful ways. This opens the door to accessible and intuitive human-robot interaction. However, free-form language often includes complex referring expressions and verbs that are difficult to map to specific objects or actions in the physical world, respectively (Lindes & Skiker, 2025).

Illustrative examples of this can be found in Human annotations datasets such as ALFRED and HandMeThat (Wan et al., 2023). In ALFRED humans observe a robot performing a task and try to write an English description of the task. In HandMeThat the robot observes a human performing a task and then tries to help based on an English language request. From these examples, human language used to describe everyday tasks is often rich in variation and ambiguity, and this are very difficult to ground. A cognitive agent that relies only on its parsing capabilities is constrained to a restricted subset of English and cannot effectively process such language. One potential solution is to integrate an LLM into its parsing strategy.

So, how can we make that possible? What solutions have been proposed so far, and what are we suggesting as a solution?

A few studies have addressed the problem of using free-form language in the context of human-robot collaboration (Gao et al., 2014; Hatori et al., 2018; Shridhar and Hsu, 2018). Hatori et al (2018) developed a robotic system which a human operator can communicate with using unconstrained spoken language instructions like *Grab that fluffy brownish stuff*. The system allows a robot to pick objects based on unconstrained human instructions and resolve ambiguities through interactive dialogue with the user. The robot can then make a decision either which object to pick and where to place it.

Our approach proposes a new way to resolve the problem of using free-form language: transforming free-form language into simplified or formal representations that the cognitive agent can interpret and act upon. To do this, we combine the reasoning capabilities of a cognitive agent with the language processing of LLMs. In our system, the LLM plays a central role in translating the human's free-form language into either a simpler form or a formal language that can fit with the agent's internal representation.

## 6.3 Experiments

This challenge of turning free-form natural language into language a cognitive agent built on the symbolic processing of Soar can understand is a major one that will require a substantial amount of research to solve. We show a simple experiment that is suggestive of what should be possible to understand a recipe obtained from ChatGPT.

---

**Request to ChatGPT:** (2/20/2025, 11:18am MST)
I need a recipe for making scrambled eggs. Please give it to me as a list of a few simple steps.

**Response from ChatGPT:**
Here's a quick and easy recipe for scrambled eggs:

**Ingredients:**
- 2–3 eggs
- Salt (to taste)
- Pepper (to taste)
- 1 tbsp butter (or oil)
- Optional: Milk or cream (1–2 tbsp for creamier eggs)

**Steps:**
1. **Crack the eggs** into a bowl and beat them with a fork or whisk until fully combined.
2. **Add a pinch of salt** and pepper to the eggs. Optionally, add a splash of milk or cream for a creamier texture.
3. **Heat a non-stick pan** over medium-low heat and add the butter.
4. **Pour the eggs into the pan** once the butter has melted, swirling the pan to coat the bottom.
5. **Cook gently**, stirring occasionally with a spatula, until the eggs are soft and slightly runny or fully cooked to your liking.
6. **Serve immediately** and enjoy your scrambled eggs!
Let me know if you'd like to adjust the recipe to your taste!

---

*Figure 9: A ChatGPT recipe for scrambled eggs*

The experiment performed here using the free ChatGPT as on 7 August 2025 is to ask the LLM to break down the procedure part of the recipe into much simpler steps. This was done using a prompt as shown in Figure 10. Notice that this prompt has no explicit examples of how inputs should produce outputs. Instead, it has an English language explanation of the desired translation, plus an inventory of components the model can use to construct its response. Those components are written with variables as arguments that the LLM is expected to understand.

```
I am a robot trying to understand human instructions. My language abilities are very
limited
I've been given a recipe for making scrambled eggs. I can sort of understand the
list of ingredients,
but the language in the steps are too complicated for me.
I'd like you to translate it into a series of short commands, one per line.
Each line should have a simple action verb followed by one or two arguments.

Here are the list of action verbs and their arguments that I can easily understand:

Pick up <object>.
Put down <object> in <container>.
Put down <object> on <surface>.
Go to <object or location>.
Turn on <instrument>.
Turn off <instrument>.
Wait <amount of time>.
Wait until <event or condition>.
Stir <substance>.
Pour <object> into <container>.
Crack <eggs> into <container>.
Serve <dish>.

Here is the procedure for making scrambled eggs:

Steps:
Crack the eggs into a bowl and beat them with a fork or whisk until fully combined.
Add a pinch of salt and pepper to the eggs. Optionally, add a splash of milk or cream
for a creamier texture.
Heat a non-stick pan over medium-low heat and add the butter.
Pour the eggs into the pan once the butter has melted, swirling the pan to coat the
bottom.
Cook gently, stirring occasionally with a spatula, until the eggs are soft and
slightly runny or fully cooked to your liking.
Serve immediately and enjoy your scrambled eggs!
Let me know if you'd like to adjust the recipe to your taste!

Please give me a list of steps to make scrambled eggs I can understand with no
additional comments:
```

*Figure 10: A prompt for explanation of a recipe*

Figure 11 shows the response that was received for the prompt in Figure 10. By and large all these instructions seem easy for our agent to understand. However, there the agent will need to use other means to learn how to perform actions like *crack* or *stir* or to recognize events like *butter is melted*. Those are not language issues, but issues for robotic perception and motor control.

```
Pick up eggs.
Go to bowl.
Crack eggs into bowl.
Pick up fork.
Stir eggs.
Put down fork on counter.
Pick up salt.
Put down salt in bowl.
Pick up pepper.
Put down pepper in bowl.
(Optional) Pick up milk.
(Optional) Pour milk into bowl.
Pick up fork.
Stir eggs.
Put down fork on counter.
Go to pan.
Turn on stove.
Wait until pan is warm.
Pick up butter.
Put down butter in pan.
Wait until butter is melted.
Pick up bowl.
Pour eggs into pan.
Put down bowl on counter.
Stir eggs.
Wait until eggs are cooked.
Turn off stove.
Serve eggs.
```

*Figure 11: A simple recipe produced by ChatGPT*

For full disclosure, the final result exactly as shown in Figure 11 was the result of the third attempt to do this task. The first attempt had two problems: ChatGPT put some general commentary before and after the list of commands, and the last command was *Serve dish.* The first problem was solved by adding *with no additional comments* to the last line of the prompt. After seeing *Serve dish.*, ChatGPT was asked *What is the dish?* It responded *The dish is **scrambled eggs**.* For the final run the line introducing the data was changed to add the word *making*. Details of this sort will probably always be a challenge when working with ChatGPT, at least as it is now.

This simple experiment looks promising as an illustration of an approach to robotic understanding of complex task descriptions provided in free-form language that is natural to humans. ChatGPT can provide general and common sense knowledge in good free-form language as in Figure 9, and it can also translate this same language into a much simpler form like that in Figure 11. Much further work is needed. As suggested by Lawley & MacClellan (2024), a strategy that breaks up the problem of understanding a text into multiple small requests will probably be necessary.

## 7. How to Get There

In this work we have discussed current commercial robot products that work closely with humans, and the fact that their collaboration with humans could be more effective if these robotic agents had good natural language understanding abilities. This is, of course, only one of several hard research problems to solve to create robots that can be useful assistants to humans in the real world, but it is an important one.

We have proposed a possible agent architecture in which a human interacts with a cognitive agent that is the center of a complete AI system consisting of that agent and the situational knowledge which is learns from experience, a physical robot that provides perception and action capabilities, and a large language model that can be a source of general and common sense knowledge and can also translate free-form human language into terms the agent can understand easily.

Within this background, we have focused on three kinds of interactions the agent might have with the LLM, and presented a simple experiment with the current ChatGPT to show the feasibility of such interactions. The first experiment shows the ability of the LLM to ground referring expressions in natural language that involve spatial relations to specific objects in the environment when given a context containing object categories and spatial neighbor relations. The second shows the ability of ChatGPT to use its general knowledge of objects and their uses to provide information about where would be likely places to store objects lying around in a kitchen. The third demonstrates that ChatGPT can take the instructions for a recipe in free-form English easy for a human to understand, but very hard for a robot, and turn it into a sequence of simple steps that fit a defined subset of simple English.

All these experiments are very simple and intended only as proofs of concept. We foresee three major challenges in using techniques like these to create a good language-capable robotic assistant.

The first problem is how the agent can break down a stream of language coming from its human collaborator or an LLM into specific, detailed questions to ask of an LLM. Our experiments show three aspects of the problem, and there are certainly more. To illustrate, consider the case of understanding a description of a complex task like the ones discussed for the third experiment. To get to the point where the agent can act correctly in the world based on a short paragraph of natural language like that in Figure 9 requires asking several specific questions of the LLM, some for clarifying free-form language, some for grounding referring expressions to objects in the immediate environment, and some for finding out how to vary the execution of a task based on the categories and properties of its arguments. It is unlikely that this can all be done correctly in a single request to the LLM, so the agent will have to know judicious ways to partition interaction for each situation.

The second problem is prompt engineering. In our simple experiments we found that it was necessary to be precise in sometimes unexpected ways to get the answers we wanted from ChatGPT. An autonomous robot might be given some innate abilities in this area, such as a number of prompt templates, but it will need to adjust things on the fly, perhaps composing a number of elements built from templates, to get responses that it can use effectively.

The third problem is the overall integration and orchestration of immediate interactions with a human, the real world, and the LLM. This will be a long-term learning enterprise. Every time the agent finds it lacks knowledge to decide what to do next, it will have to decide how to combine the resources in its long-term situational knowledge, requests of the LLM or other external data sources, and interactions with its human collaborator. We believe this is the biggest challenge for "agentic LLMs," and that the several decades of research experience with cognitive architectures and cognitive agents is the best place to look to find solutions.

## Acknowledgments

The authors wrote this chapter as unfunded independent research. Many of the ideas were inspired by our work with our respective organizations. James Kirk's experiments on grounding object references using an LLM was the inspiration for designing the experiments described here. John Laird, Aaron Mininger, James Kirk, Bob Wray, and Steven Jones at the Center for Integrated Cognition have been instrumental in developing many of the ideas on cognitive agents and their use with LLMs that have contributed to what is presented here.